\documentclass{article}

\PassOptionsToPackage{numbers, compress}{natbib}
\usepackage[preprint]{neurips_2024}

\usepackage[utf8]{inputenc} 
\usepackage[T1]{fontenc}    
\usepackage{hyperref}       
\usepackage{url}            
\usepackage{booktabs}       
\usepackage{amsfonts}       
\usepackage{nicefrac}       
\usepackage{microtype}      
\usepackage{xcolor}         

\usepackage{amssymb}
\usepackage{bm}
\usepackage{amsmath}
\usepackage[ruled,linesnumbered]{algorithm2e}
\usepackage{color}
\usepackage{array}
\usepackage{multirow}
\usepackage{ragged2e}
\usepackage{tabularray}
\usepackage{tikz}
\usepackage{pgfplots}
\usepackage{ulem} 
\usepackage{graphicx}
\usepackage{subfig}
\usepackage{float}

\title{Rethinking Class-Incremental Learning from a Dynamic Imbalanced Learning Perspective}

\author{
  Leyuan Wang \\
  Beijing University of \\Posts and Telecommunications \\
  \texttt{leyuan.wang@bupt.edu.cn} \\
  \And
  Liuyu Xiang\thanks{Corresponding author.} \\
  Beijing University of \\Posts and Telecommunications \\
  \texttt{xiangly@bupt.edu.cn} \\
  \And
  Yunlong Wang\\
  Institute of Automation,\\ 
  Chinese Academy of Sciences \\
  \texttt{yunlong.wang@cripac.ia.ac.cn} \\
  \And
  Huijia Wu \\
  Beijing University of \\Posts and Telecommunications \\
  \texttt{huijiawu@bupt.edu.cn} \\
  \And
  Zhaofeng He \\
  Beijing University of \\Posts and Telecommunications \\
  \texttt{zhaofenghe@bupt.edu.cn} \\
}

\begin{document}

\maketitle

\begin{abstract}
Deep neural networks suffer from catastrophic forgetting when continually learning new concepts. In this paper, we analyze this problem from a data imbalance point of view. We argue that the imbalance between old task and new task data contributes to forgetting of the old tasks. Moreover, the increasing imbalance ratio during incremental learning further aggravates the problem. To address the dynamic imbalance issue, we propose Uniform Prototype Contrastive Learning (UPCL), where uniform and compact features are learned. Specifically, we generate a set of non-learnable uniform prototypes before each task starts. Then we assign these uniform prototypes to each class and guide the feature learning through prototype contrastive learning. We also dynamically adjust the relative margin between old and new classes so that the feature distribution will be maintained balanced and compact. Finally, we demonstrate through extensive experiments that the proposed method achieves state-of-the-art performance on several benchmark datasets including CIFAR100, ImageNet100 and TinyImageNet.
\end{abstract}

\section{Introduction}
\label{sec:intro}

Humans possess a strong adaptability to handle external changes and are able to continuously acquire, update, accumulate and utilize knowledge. Naturally, we expect Artificial Intelligence (AI) systems to adapt in a similar manner, i.e., incremental learning or lifelong learning. Taking the most fundamental classification scenario as an example, Class-Incremental Learning (CIL) aims to construct a holistic classification model among all seen classes across multiple tasks. However, when the model is trained on new class data, its performance on old classes tends to degrade drastically, which is also known as \textit{catastrophic forgetting} \cite{mccloskey1989catastrophic,goodfellow2013empirical}. A major dilemma faced by incremental learning is the trade-off between plasticity and stability.

Numerous efforts have been devoted to addressing the above challenges, such as retaining a small amount of old data and replaying in new tasks \cite{rebuffi2017icarl,castro2018end,chaudhry2018riemannian}, regularization on the network parameter changes \cite{kirkpatrick2017overcoming,zenke2017continual}, maintaining knowledge \cite{LiH16,rebuffi2017icarl} from previous tasks, and dynamically expanding the model structure in new tasks \cite{yan2021dynamically,WangZYZ22,D0001WYZ23}. 
Among them, replay-based methods, which typically pre-allocate a fixed-size memory and store subsets of old class data as exemplars for experience replay, have demonstrated promising performance. The strategy of reviewing past knowledge aligns with the principles of human cognition and has achieved desirable performance in previous methods.

\begin{figure}[!htb]
\centering
\includegraphics[width=0.6\textwidth]{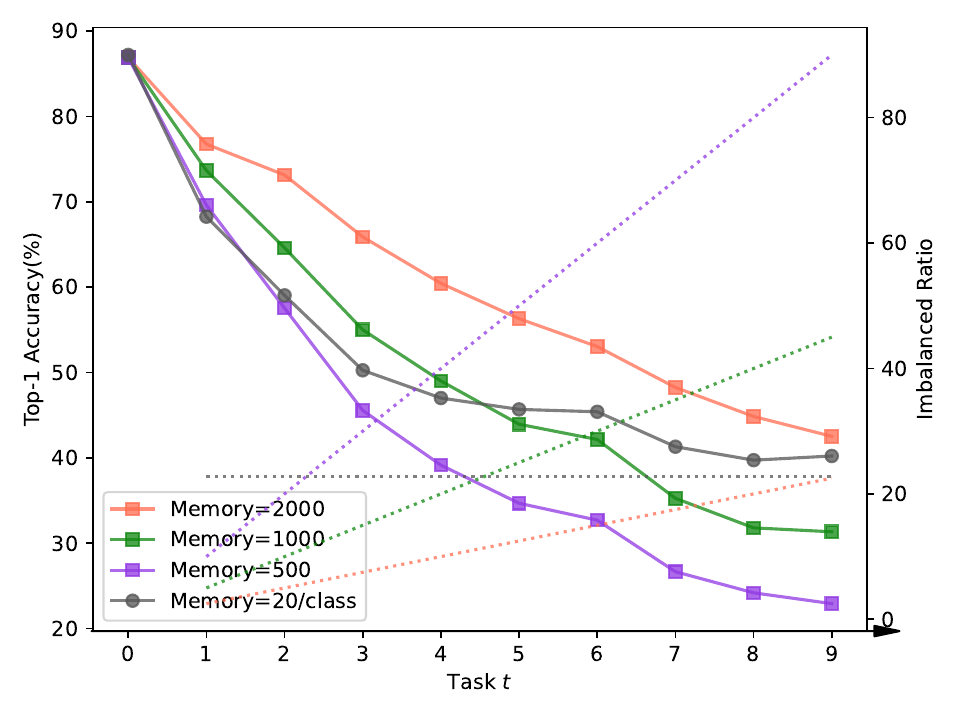}\label{subfig:memory_size_imblance}
\caption{
The imbalance ratio increase dynamically in CIL and is negatively correlated with memory size. The solid line represents the accuracy on each task, while the dashed line represents the imbalance ratio. The above experiments are performed by iCaRL on CIFAR100.}
\label{fig:imblance}
\end{figure}

However, due to storage or privacy constraints, the memory capacity is very limited, and the number of old class exemplars is far less than that of new class samples. The replay-based CIL methods have two memory management strategies: the first strategy is storing a fixed number of samples per class \cite{hou2019learning,LiuSS21,abs-2308-01698}, while the second approach pre-allocates a fixed capacity of memory for all seen class \cite{rebuffi2017icarl,abs-2308-01698}. Unlike classical imbalance learning\cite{LinGGHD17,MenonJRJVK21,KangLXYF21,RenYSMZYL20,FengZH21}, CIL is a dynamic process. For the latter memory management strategy, the degree of imbalance intensifies with the increase in classes, further damaging the performance of CIL. Here, we introduce the imbalance ratio (IR) to measure the degree of imbalance in the training data, which is defined as the ratio of the sample size of the largest majority class to that of the smallest minority class. As shown in Fig.~\ref{fig:imblance}, in each training task, the accuracy drops as the IR increases, indicating a more negative impact of the imbalance. Furthermore, the fixed total memory method performs better when the IR is smaller than the class-wise fixed memory method, whereas its performance is worse. Therefore, in this paper, we focus on tackling the dynamic imbalance problem in memory-based CIL. 

Intuitively, the data imbalance further causes the decision boundary between the minority (old) class and the majority (new) class to be biased, leading to performance degradation on old classes. As shown in Fig.~\ref{fig:feature_space}(a), in task $T$, suppose the neural network has a balanced feature space using softmax cross-entropy loss, with no margin between classes. At the beginning of task $T+1$, new classifiers are randomly initialized and fall within the decision boundary of the old class, as shown in Fig.~\ref{fig:feature_space}(b). During the training of task $T+1$, all the old class samples are sampled from the memory and are far fewer than those of the new class. The softmax cross-entropy loss results in a biased classifier, which compresses the decision boundary of the old class, as shown in Fig.~\ref{fig:feature_space}(c). There are also empirical findings \cite{hou2019learning,ZhaoXGZX20} showing that the norm of the classifier corresponding to the minority (old) class and the majority (new) class is imbalanced during the incremental learning, indicating a biased decision boundary. Inspired by the previous study \cite{kang2019decoupling} showing that representation learning matters in the imbalance learning, we consider the data imbalance problem in CIL from the perspective of representation learning and raise the following question:

\begin{figure}[t]
\centering 
    \centering 
    \includegraphics[width=\textwidth]{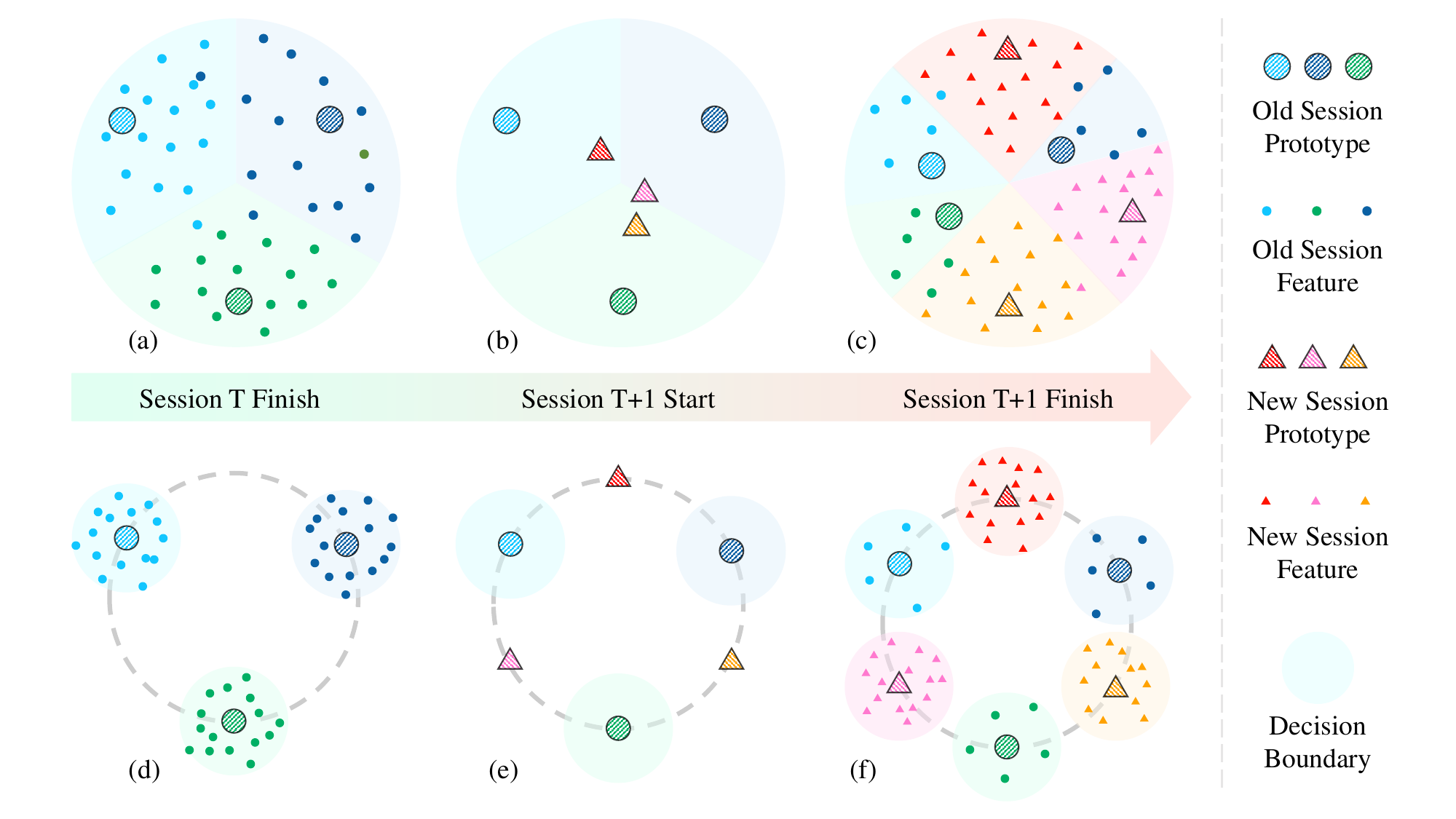} 
    \caption{Comparison between the imbalanced feature space (a-c) learned by classic incremental learning methods and the uniform feature space (d-f) learned by the proposed UPCL.} 
    \label{fig:feature_space} 
\end{figure}

\noindent\textbf{What is the most appropriate feature space for incremental learning?}

We argue that an appropriate feature space for incremental learning can maintain the feature distribution of old classes while adapting to the feature distribution of new classes. It should have the following characteristics: 1) \textbf{class uniformity}: the intra-class feature should be uniformly distributed in the feature space to avoid confusion, as shown in Fig.~\ref{fig:feature_space}(d). 2) \textbf{feature compactness}: the intra-class feature distance should be much smaller than the inter-class feature distance. Considering the dynamic process of CIL, space should be reserved for new classes that may appear in the future, as shown in Fig.~\ref{fig:feature_space}(e)-\ref{fig:feature_space}(f).

Based on the above analysis, we propose uniform prototype contrastive learning (UPCL). First, we define a set of non-learnable prototype uniformly distributed on a unit-hypersphere before the task starts, and use a heuristic method to assign prototypes to classes, thus ensuring class uniformity. Then, we employ contrastive learning to make the features as close as possible to their corresponding prototype while moving away from prototypes of other classes. Furthermore, we dynamically adjust the relative margin between new and old class features in the contrastive learning with the statistical prior of the training data. In this way, we mitigate the data imbalance in each task under the guidance of uniform prototypes.

To verify the effectiveness of the proposed method, we conduct extensive experiments on benchmark datasets including CIFAR100, ImageNet100 and TinyImageNet. We demonstrate that the proposed UPCL method outperforms state-of-the-art methods in standard CIL. In conclusion, our main contributions are summarized as follows:

\begin{itemize}
\item We investigate the data imbalance problem in CIL and analyze that a feature space with class uniformity and feature compactness would be beneficial for CIL.
\item We propose uniform prototype contrastive learning (UPCL), where a set of pre-defined uniform prototypes are designed to guide the feature learning to achieve class uniformity. We further propose prototype contrastive learning with dynamic margin to learn a compact feature distribution. 
\item We conduct comprehensive experiments on several benchmark datasets to verify the effectiveness of UPCL. The results show that UPCL achieves state-of-the-art performance in standard CIL settings.
\end{itemize}

\section{Related Work}

\noindent \textbf{Class-Incremental Learning.} Various methods have been proposed to analyze the causes of catastrophic forgetting and tried to alleviate this phenomenon. Existing methods can be coarsely categorized into three types: model-based, data-based, and algorithm-based methods \cite{abs-2302-03648}.  

Model-based methods \cite{kirkpatrick2017overcoming,zenke2017continual,yan2021dynamically,WangZYZ22,D0001WYZ23} focus on adapting the model parameters or architecture to accommodate new tasks without significantly affecting the performance on previous tasks. 

Data-based methods focus on storing an additional set of exemplars for data replay during the learning of new tasks \cite{rebuffi2017icarl,castro2018end,chaudhry2018riemannian,abs-2308-01698}. 
The strategy of reviewing past knowledge aligns with the principles of human cognition and has achieved desirable performance in previous experiments, hence it is widely applied in various continual learning baselines. However, since the memory only holds a small portion of the training set, data replay methods suffer from over-fitting and data imbalance. 

Algorithm-based methods concentrate on designing algorithms that can effectively maintain the knowledge from previous tasks, where the knowledge distillation technique \cite{hinton2015distilling} is commonly used to establish a mapping between the old and new models. For instance, LwF \cite{LiH16} preserves an additional old model to constrain the new model's output on the old classes. Model rectification rectifies the anomalous behavior of continual learning models. LUCIR \cite{hou2019learning} maximizes the inter-class distance in the cosine space and uses feature distillation to prevent forgetting. WA \cite{ZhaoXGZX20} suggests post-hoc normalization of the weights to ensure that the predicted probability is proportional to the classifier weights. 
Building upon WA, the CwD \cite{shi2022mimicking} is regularized the feature representation to make it more uniform. NCCIL \cite{yang2023neural} introduces the concept of Neural Collapse \cite{papyan2020prevalence} and uses regression to make the feature representation converge towards a fixed classifier.

These methods empirically observe the impact of imbalance on continual learning, such as the imbalance of the logit model and the confusion between new and old classes, and propose some solutions from a static perspective. However, these methods do not explicitly realize the differences between the two types of memory management strategies, making it difficult to analyze from the perspective of dynamic imbalance learning and have not effectively utilized the prior knowledge of imbalance.

\noindent \textbf{Imbalanced Learning.} Data in the real world typically follows an imbalanced distribution, which causes network training to be biased towards dominant classes, leading to a performance drop. Common solutions involve re-sampling \cite{wang2014resampling,chang2021image,wei2022open} and designing re-weighting loss functions \cite{LinGGHD17,RenYSMZYL20,FengZH21}.
Recent work also finds that decoupling representation learning from classifier learning can yield more distinctive feature representations, thereby enhancing the generalizability of the model \cite{kang2019decoupling}. 
TSC\cite{li2022Targeted} uses unlearnable prototypes that are pre-generated and approximately uniformly distributed on a hypersphere as targets for supervised contrastive learning. This approach improves the uniformity of the feature distribution on the hypersphere and achieves state-of-the-art performance on long-tailed recognition sessions. The change in the number of classes and training data distribution make it difficult to directly apply imbalance learning methods in CIL.

\section{Methodology}

\begin{figure*}[!htb]
\centering 
\includegraphics[width=\textwidth]{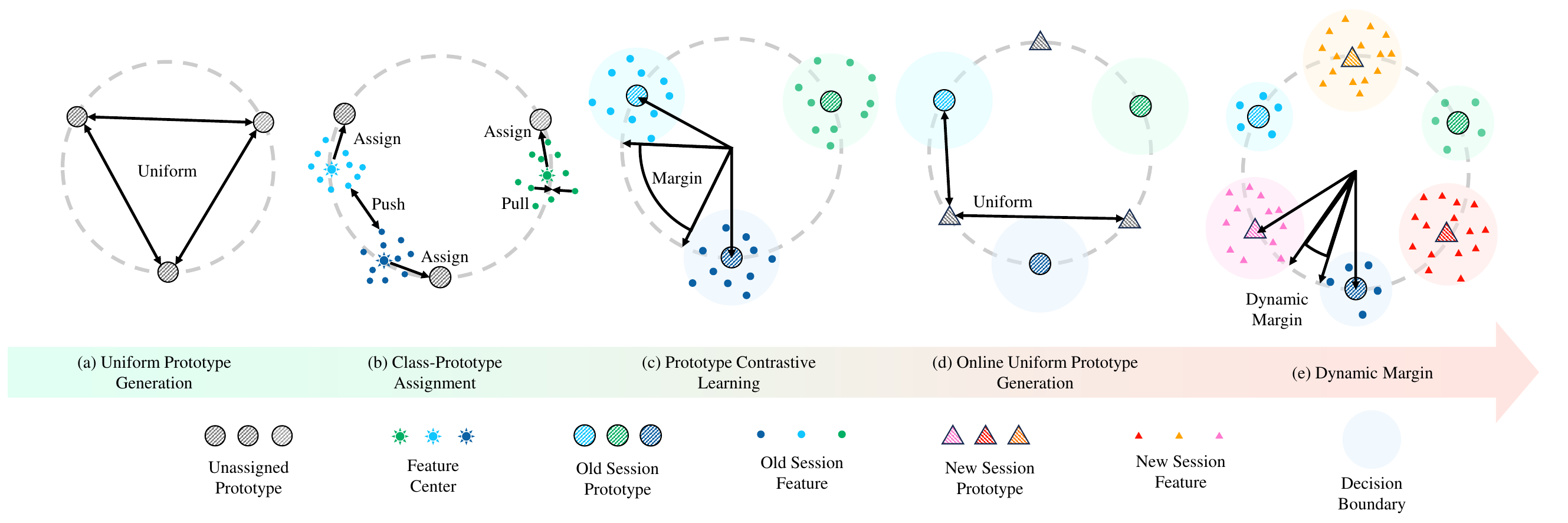} 
\caption{The pipeline of uniform prototype contrastive learning (UPCL). Our method can dynamically generate uniform prototypes on the unit-hypersphere during the CIL process, and heuristically assign classes to the prototypes during the training process, thus ensuring uniformity. Prototype contrastive learning with dynamic margin dynamically adjusts the relative margin between new and old class features in the contrastive loss using the statistical prior of the training data, thus ensuring compactness.} 
\label{fig:framework} 
\end{figure*}

In this paper, we propose uniform prototype contrastive learning (UPCL). It uses pre-defined uniform prototypes to guide feature learning and adopts contrastive learning with dynamic margin to learn compact representations. The pipeline of UPCL is shown in Fig.~\ref{fig:framework}.


\subsection{Problem Formulation}

In class-incremental learning (CIL) scenario, the model needs to learn a unified classifier that can classify all classes that have been learned at different stages. Specifically, $\mathcal{D}=\{\mathcal{D}_t\}_{i=1}^T$ is a sequence training tasks, where $\mathcal{D}_t=\{(\mathbf{X}_t,\mathbf{Y}_t)\}=\{(x^t_i,y^t_i)\}_{i=1}^{N_t}$ is the $t$-th incremental task and $x^t_i \in \mathcal{R}$ is a training sample of class $y^t \in \mathbf{Y}_t$. $\mathbf{Y}_t$ is the label set of task $t$, where $\mathbf{Y}_t \cap \mathbf{Y}_{t'} = \emptyset$ for $t \neq t'$.  When training on task $t$, only $\mathcal{D}_t$ and memory $\mathcal{M}$ can be accessed, where the data in $\mathcal{M}$ are sampled from $\{\mathcal{D}_1,\mathcal{D}_2,...,\mathcal{D}_{t-1}\}$ and its number $N_\mathcal{M}$ is limited such that $N_\mathcal{M} \ll \sum_{i=0}^{T-1}{N_i}$. The model consists of a representation network $f(x;\theta_t)$ and a unified classifier.

\subsection{Uniform Prototype Generation}

Since we have limited access to the old task data, the task-level data imbalance problem arises during the new task learning stage, where the minority classes (old classes) are suppressed by the majority classes (new classes).
From the perspective of feature representation, this manifests as feature collapse, where the feature representations of minority classes drift towards majority classes and become confused. Therefore, it is necessary to learn distinguishable feature representations to prevent such confusion. 

Following the spirit, we generate a set of unlearnable class prototypes before new task training. These prototypes $\bm{\phi}$  are uniformly distributed on the unit-hypersphere $S^{d-1}$, and the features are made to converge to the prototypes during the training process, so that clear decision boundaries and distinguishable feature representations can be obtained. To this end, we initialize new task prototypes $\bm{\phi}_t$ before the start of task $t$ with the Gram-Schmidt algorithm \cite{cheney2009linear}.

For simplicity, we consider a single-layer linear classifier without bias, whose weights can be seen as class prototypes $\bm{\phi}_t$. We normalize the extracted feature and the prototypes so that they are distributed on a unit-hypersphere $S^{d-1}$. In task $t$, the goal is to minimize a predefined loss function $\mathcal{L}(\mathbf{X}_t, \mathbf{Y}_t;\theta_t,\bm{\phi}_t)$ on $\mathcal{D}_t$ and $\mathcal{M}$ while preserving the knowledge learned from previous task: 
\begin{equation}
{\theta_t,\bm{\phi}_t} = {\arg\min} \ \mathcal{L}(f(\mathbf{X}_t;\theta_t)\odot \bm{\phi}_t, \mathbf{Y}_t).
\end{equation}

Note that we use orthogonality instead of strict uniformity, because it can be considered as an approximation of uniformity. Here we adopt the definition of uniformity in the Tammes problem\cite{tammes1930origin}, that is, the maximization of the minimum distance between points on the surface. When the number of classes or prototypes is less than the dimensionality, which is the case for most CIL tasks \footnote{The scenario where the total number of classes exceeds the feature dimension is not considered in this paper, and it due to the insufficient representation ability. A more reasonable approach would be to increase the network capacity and the feature dimension in CIL, but this is beyond the scope of this paper.}, according to \cite{graf2021dissecting,papyan2020prevalence}, the distribution of uniform points constitutes the simplex equiangular tight frame (i.e $\forall i,j,i \ne j,\sum_i{t_i=0},\exists \Delta \in \mathbb{R} , t^T_i \cdot  t_j = \Delta$). But simplex equiangular tight frame is difficult to construct dynamically and thus not suitable for incremental learning scenarios.  While orthogonality ensures that the cosine distance between any two prototypes is at least 1, and the Gram-Schmidt algorithm itself has the property of online computation without imposing a significant computational burden. In supplementary material, we validate that orthogonality indeed approximates uniformity by comparing the minimum cosine distances of prototypes generated by several different methods.


\subsection{Class-Prototype Assignment}

Unlike learnable prototypes, unlearnable prototypes require assigning a class to each prototype. Ideally, classes that are semantically close to each other should be assigned to similarly close positions to ensure consistency. However, it is difficult to accurately quantify the degree of semantic proximity between two classes. To this end, a modified Hungarian algorithm \cite{li2022Targeted} is applied to find the good assignment $\bm{\sigma}$ for the new classes. The algorithm aims to minimize the distance between the feature centers $\mu_t$ of the new classes and their prototypes $\bm{\phi}_t$:
\begin{equation}
\bm{\sigma} = \arg \min \frac{1}{C}\sum^{C}_{i=1}||\phi^i_t -\mu^i_t||,
\label{func:targetmatch}
\end{equation}
where $C_t$ is the number of new classes, and $\mu_t \leftarrow \lambda \cdot \mu_t+(1-\lambda) \cdot \mu_t'$  is computed with the exponential moving average during training. As the training progresses, the distance between feature centers $\mu_t$ and their corresponding prototypes $\bm{\phi}_t$ decreases. In our experiments, it is observed that the assignment stabilizes after a few epochs, so the assignment algorithm does not impose much computational burden.

\subsection{Prototype Contrastive Learning with Dynamic Margin}

To learn discriminative features from the assigned uniform prototypes, we propose the uniform prototype contrastive loss $\mathcal{L}_{\text{UPCL}}$, which is inspired by InfoNCE \cite{He0WXG20} and AM-softmax\cite{0015LDL018}. 
Specifically, on the one hand, since the unassigned prototypes are already uniformly distributed in the feature space, we hope that similar features are close to each other to assign class-prototype. On the other hand, we hope that the features are close enough to the corresponding prototype and maintain a sufficient margin with other classes to avoid confusion and reserve space for future classes.

Given a batch of features $V_t = \{v_0,v_1,...v_i\}$ in task $t$, $U_t = V_t \cup \bm{\phi}^t$, and $U^{i}_t \subset U_t$ is a positive set containing features selected from the same class as $v_i$ and its corresponding prototype. $\mathcal{L}_{\text{UPCL}}$ is defined as:
\begin{equation}
\mathcal{L}_{\text{UPCL}} = - \sum_{\substack{v_i \in V_t,\\ u^{i}_j \in U^{i}_t}} \log \frac{\text{exp}(\psi(v_i,u^{i}_j) / \tau)}{\sum_{u_k \in U_t} \text{exp}(\psi(v_i , u_k) / \tau)},
\label{func:UPCL}
\end{equation}
\noindent where $\tau$ is a temperature hyper-parameter that can adjust the convergence speed and generalization. $\psi(\cdot)$ is a metric function. Considering that $v_i$ and $\phi_i$ are vectors on the unit-hypersphere $S^{d-1}$, their inner product is the cosine similarity. For features $v_i$, $v_j$ and prototype $\phi$, we have

\begin{equation}
\psi(v_i,v_j)=v_i \cdot v_j, \quad
\psi(v_i,\phi)=v_i \cdot \phi - m,
\end{equation}

\noindent where $m$ is an additive margin. To address the issue of imbalanced data, we aim to provide a larger additive margin for minority classes to avoid confusion. Due to the fact that the number of examples in the old class is much smaller than that of the new class, it is more likely to be misclassified as a hard sample of the new class. Therefore, according to Menon et.al \cite{MenonJRJVK21} who propose an elegant solution from a statistical perspective, $m =- \log p(y)$, which $p(y)$ is the prior distribution. Specifically, we count the frequency of sample classes in the training data as $p(y)$ before the start of each task. And by introducing prior distribution, the loss function can fit the mutual information between the prototype and the feature in a stable convergent way, thus learning the truly important correlations.

From the perspective of $U_t = V_t \cup \bm{\phi}^t$, we can decompose $\mathcal{L}_{\text{UPCL}}$ into two parts: $\mathcal{L}_{\text{proto}}$ and $\mathcal{L}_{\text{feat}}$.

\begin{equation}
\mathcal{L}_{\text{proto}} = - \sum_{\substack{v_i \in V_t, \\ \phi_k \in \bm{\phi}}} \log \frac{\text{exp}((v_i \cdot \phi_k + \log p(y))  / \tau)}{\sum_{\phi_k \in \bm{\phi}} \text{exp}((v_i \cdot \phi_k + \log p(y)) / \tau)},
\label{func:PSC}
\end{equation}

\begin{equation}
\mathcal{L}_{\text{feat}} = - \sum_{\substack{v_i \in V_t,\\v_j \in V_i}}\! \log \frac{\text{exp}(v_i \cdot v^{+}_j/\tau)}{\sum_{v_j \in V} \text{exp}(v_i \cdot v_j/ \tau)},
\end{equation}

\noindent where $v^{+}_i\subset V_t$ is a positive set containing features selected from the same class as $v_i$. 
$\mathcal{L}_{\text{feat}} $ is a standard supervised contrastive loss \cite{KhoslaTWSTIMLK20}, which can intuitively pull together features of the same class and push away features of different classes. $\mathcal{L}_{\text{proto}}$ is a softmax cross-entropy loss with temperature and dynamic margin, which can intuitively make features close to their corresponding prototypes and keep a certain distance from other prototypes. The prior distribution acts as a margin that allows minority classes to maintain a larger distance and avoid confusion. 

Finally, our loss function is defined as $\mathcal{L}_{\text{UPCL}} = \mathcal{L}_{\text{proto}} + \lambda^t \mathcal{L}_{\text{feat}}, $
\noindent where $t=0,1,...T$ and $\lambda$ is a decaying factor. We noticed in experiments that $\mathcal{L}_{\text{feat}}$ helps the model to learn distinctive features in the initial tasks. However, in subsequent tasks, it causes the features to be overly uniform, lacking compactness. This also indicates the crucial importance of dynamic margin. Therefore, we empirically set $\lambda=0.5$ to dynamically reweight $\mathcal{L}_{\text{feat}}$.

In this way, we use uniform prototypes to guide the model to learn a uniform class distribution, while dynamic margin make the features compact. This effectively addresses the imbalance problem in CIL and reduces confusion between new and old classes.

\section{Experiments and Results}

\subsection{Datasets and Evaluation Metrics}

We conduct comprehensive experiments on several benchmark datasets, including CIFAR100, ImageNet100 and TinyImageNet. CIFAR100 \cite{krizhevsky2009learning} is a relatively small dataset that consists of 50,000 training images with 500 images per class, and 10,000 testing images with 100 images per class. The images in CIFAR100 are RGB with a size of $32 \times 32$ pixels. ImageNet100 is a subset of 100 classes randomly selected from ImageNet~\cite{deng2009imagenet}, which consists of 129,395 training images and 5,000 testing images. The images in ImageNet100 are RGB with a size of $224 \times 224$ pixels. TinyImageNet consists of 100,000 training images with 500 images per class, and 10,000 testing images with 50 images per class. All images in TinyImageNet are downsampled to $64 \times 64$. For CIL, following the setting in iCaRL \cite{rebuffi2017icarl}, all the classes are arranged in a fixed random order and then equally splitted into a sequence of tasks.

We record the top-1 accuracy $\mathcal{A}$ after task $t$ and present it as a curve, using the accuracy at the end of the last task as a metric for overall accuracy, denoted as $\mathcal{A}_{\text{last}}$. In addition, the average accuracy considers the performance of all tasks, denoted as $\mathcal{A}_{\text{avg}} = \frac{1}{T} \sum_{t=0}^{T-1} \mathcal{A}_t,$
where $T$ is the number of tasks.

\subsection{Baseline and Implementation Details}

To highlight the effectiveness of our proposed method, we adopted a simple pipeline similar to iCaRL \cite{rebuffi2017icarl}. Without bells and whistles, our method did not adopt dynamic networks, advanced rehearsal strategies, self-supervised learning auxiliary tasks, or other hyper-parameter tuning tricks. 

To ensure a fair comparison with other methods, we implemented UPCL on the PyCIL framework \cite{zhou2023pycil}. We compared iCaRL \cite{rebuffi2017icarl}, LwF \cite{LiH16}, WA \cite{ZhaoXGZX20}, FeTrIL\cite{petit2023fetril} and NCCIL \cite{yang2023neural}, which are also implemented on PyCIL and use the same hyper-parameters. We also combine iCaRL with the three classic imbalance learning methods, FocalLoss \cite{LinGGHD17}, LA \cite{MenonJRJVK21}, and LDAM \cite{cao2019learning}, and implement them using PyCIL.
Additionally, we also referenced the performance of Coil \cite{zhou2021co} and BiC \cite{WuCWYLGF19}, and replicated the results of LUCIR \cite{hou2019learning}, CwD \cite{shi2022mimicking}, ScrollNet\cite{yang2023scrollnet} using the official code and the original settings.
In all experiments, the above methods used ResNet18 as the backbone network. 
All methods used the same rehearsal strategy as iCaRL, and the memory size is 2000 unless otherwise specified.

\begin{table}
\caption{\textbf{Results on CIFAR100, Imagenet100 and TinyImagenet.} \textit{Oracle} and \textit{Finetune} respectively represent joint training on all data and only training on the current task data, which are the upper and lower boundaries of CIL performance.}
\label{tab:main_result}
\centering
\scalebox{0.85}{
\begin{tblr}{
  width = \linewidth,
  colspec = {Q[125]Q[79]Q[79]Q[79]Q[79]Q[79]Q[79]Q[81]Q[81]Q[84]Q[84]},
  cells = {c},
  cell{1}{1} = {r=3}{},
  cell{1}{2} = {c=6}{0.474\linewidth},
  cell{1}{8} = {c=2}{0.162\linewidth},
  cell{1}{10} = {c=2}{0.168\linewidth},
  cell{2}{2} = {c=2}{0.158\linewidth},
  cell{2}{4} = {c=2}{0.158\linewidth},
  cell{2}{6} = {c=2}{0.158\linewidth},
  cell{2}{8} = {c=2}{0.162\linewidth},
  cell{2}{10} = {c=2}{0.168\linewidth},
  hline{1,20} = {-}{0.1em},
  hline{2} = {2-7}{leftpos = -1, rightpos = -1, endpos},
  hline{2} = {8-9}{leftpos = -1, rightpos = -1, endpos},
  hline{2} = {10-11}{leftpos = -1, rightpos = -1, endpos},
  hline{3} = {2-3}{leftpos = -1, rightpos = -1, endpos},
  hline{3} = {4-5}{leftpos = -1, rightpos = -1, endpos},
  hline{3} = {6-7}{leftpos = -1, rightpos = -1, endpos},
  hline{3} = {8-9}{leftpos = -1, rightpos = -1, endpos},
  hline{3} = {10-11}{leftpos = -1, rightpos = -1, endpos},
  hline{4} = {-}{},
  hline{6} = {-}{dashed},
  hline{9} = {-}{dashed},
  hline{11} = {-}{dashed},
}
         & CIFAR100      &                &                &                &                &                & Imagenet100    &                & TinyImagenet   &                \\
         & T=5            &                & T=10           &                & T=20           &                & T=20           &                & T=40           &                \\
         & $\mathcal{A}_\text{last}$           & $\mathcal{A}_\text{avg}$            & $\mathcal{A}_\text{last}$            & $\mathcal{A}_\text{avg}$            & $\mathcal{A}_\text{last}$            & $\mathcal{A}_\text{avg}$            & $\mathcal{A}_\text{last}$            & $\mathcal{A}_\text{avg}$            & $\mathcal{A}_\text{last}$            & $\mathcal{A}_\text{avg}$            \\
\textit{Oracle}        & 76.76          & -              & 76.76          & -              & 76.76          & -              & 79.12          & -              & 41.45          & -              \\
\textit{Finetune}      & 17.40          & 38.38          & 8.98           & 26.16          & 5.00           & 17.44          & 4.72           & 16.77          & 2.60           & 9.70           \\
FocalLoss     & 42.49              & 61.19          & 36.99              & 56.02          & 32.97              & 55.00          & 15.46              & 46.52              & 3.02              & 22.98          \\
LDAM     & 50.66              & 66.61          & 40.28              & 55.48          & 41.87              & 60.89          & 37.9              & 58.21              & 10.43              & 28.58          \\
LA     & 49.50              & 65.18          & 42.80              & 58.86          & 40.82              & 59.21          & 38.4              & 58.18              & 11.21              & 29.45          \\
LwF           & 34.42          & 51.84          & 24.72          & 44.58          & 14.30          & 33.51          & 15.56          & 38.47          & 7.25           & 20.13          \\
FeTrIL        & 45.13          & 60.73          & 33.93          & 50.18          & 27.40          & 43.76          & 6.84           & 18.38          & 4.06           & 9.67           \\
iCaRL         & 47.48          & 63.72          & 42.51          & 61.01          & 41.59          & 60.38          & 36.42          & 56.94          & 16.62          & 35.01          \\
LUCIR     & 41.50              & 55.67          & 31.60              & 51.43          & 28.60              & 49.27          & -              & -              & -              & -          \\
Coil          & 45.54          & 63.33          & 39.85          & 60.27          & 34.33          & 57.68          & 34.00          & 56.21          & -              & -              \\
BiC           & 56.22          & 67.03          & 50.79          & 65.08          & 43.08          & 62.38          & 34.56          & 58.03          & -              & -              \\
WA            & 60.57          & 70.43          & 55.91          & 68.22          & 45.71          & 63.28          & 46.96          & 65.35          & 13.49          & 32.90          \\
CwD     & 40.10              & 54.98          & 31.50              & 51.17          & 25.70              & 47.65          & -              & -              & -              & -          \\
NCCIL         & 53.65          & 61.09          & 48.53          & 60.93          & 42.84          & 60.08          & 44.53          & 61.61          & 15.87          & 34.64 \\
ScrollNet     & 42.30              & 60.46          & 32.79              & 54.52          & 25.88              & 49.37          & -              & -              & -              & -          \\
\textbf{UPCL}     & \textbf{60.92} & \textbf{70.82} & \textbf{56.12} & \textbf{69.41} & \textbf{51.30} & \textbf{67.16} & \textbf{48.88} & \textbf{67.50} & \textbf{19.65} & \textbf{37.31} 
\end{tblr}}
\end{table}

\subsection{Quantitative Results}

\noindent \textbf{Results on CIFAR100.} The results on CIFAR100 are shown in Table~\ref{tab:main_result}. The table includes the last accuracy and average accuracy of each method. We evaluate three settings with T=5, 10, and 20, where each task has 20, 10, 5 new classes respectively. It can be observed that UPCL outperforms all the other CIL methods in all settings and metrics. UPCL achieves an $\mathcal{A}_\text{avg}$ of 70.82\%, 69.41\%, and 67.16\% for T=5, 10, and 20 respectively, which are 0.97\%, 1.19\%, and 3.88\% higher than the second-best method WA. This demonstrates the effectiveness of UPCL in a progressive manner. UPCL also shows a consistent improvement in the $\mathcal{A}_\text{last}$, which indicates that UPCL can better preserve the knowledge of previous classes and reduce the forgetting problem. Meanwhile, we observed that when T=5, both LDAM and LA, two classic imbalanced learning methods, exhibit some performance improvement compared to the baseline iCaRL. However, as the number of tasks increases, the performance of all three classic imbalanced learning methods actually falls below the baseline. This suggests that applying classic imbalanced learning methods directly to CIL is unreasonable.

\noindent \textbf{Results on ImageNet100 and TinyImagenet.} Due to space constraints, only the results for T=20 on Imagenet100 and T=40 on TinyImagenet (each task has 5 new classes for both datasets) are shown in Table~\ref{tab:main_result}, with more detailed results to be presented in the supplementary material. The results show that our proposed UPCL outperforms other methods in terms of both $\mathcal{A}_\text{last}$ and $\mathcal{A}_\text{avg}$. UPCL achieves the highest $\mathcal{A}_\text{last}$ of 48.88\% and the highest $\mathcal{A}_\text{avg}$ of 67.50\% on ImageNet100. On TinyImagenet, UPCL achieves the highest $\mathcal{A}_\text{last}$ of 19.65\% and the highest $\mathcal{A}_\text{avg}$ of 37.31\%. Unlike CIFAR100, TinyImagenet has   more tasks in the shown experiment and more classes, which poses greater challenges. Nevertheless, our method still performs well. In summary, our proposed UPCL consistently outperforms all comparative methods across all experiments, exhibiting a substantial margin over the second-best method.

\begin{figure*}[!ht]
\centering
    \subfloat[ImageNet100 $T=5$.]{\includegraphics[width=.45\linewidth]{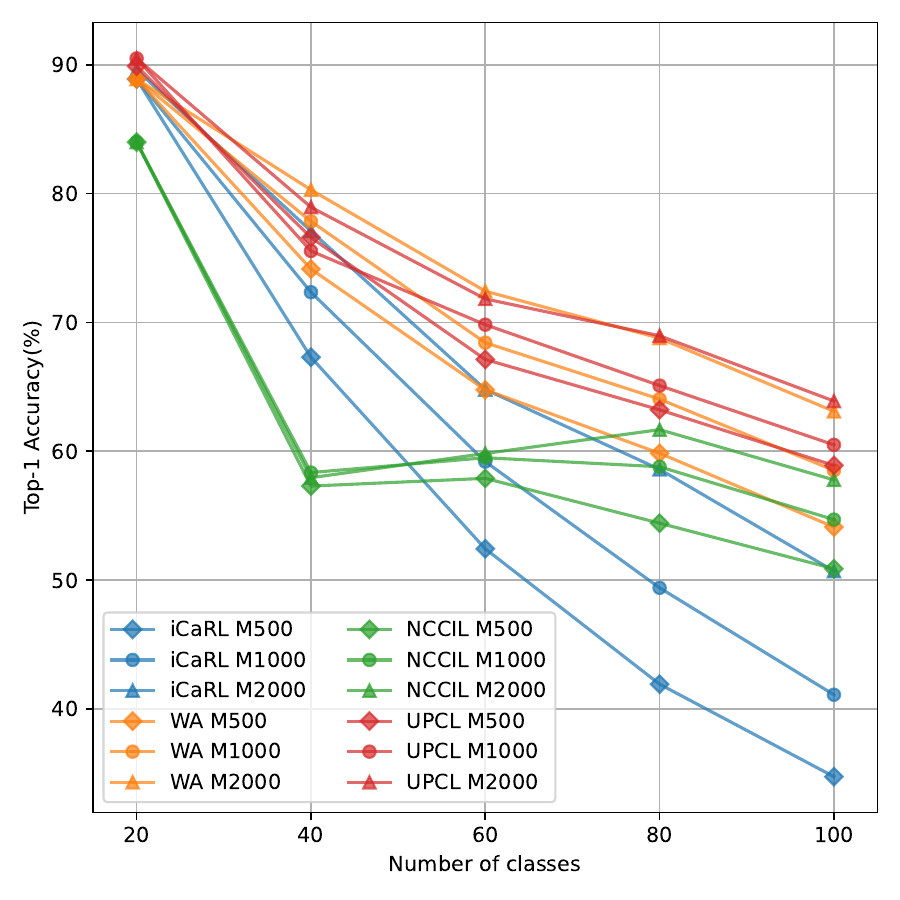}\label{subfig:imagenet100_0_20}}
    \hspace{5pt}
    \subfloat[TinyImageNet $T=20$.]{\includegraphics[width=.45\linewidth]{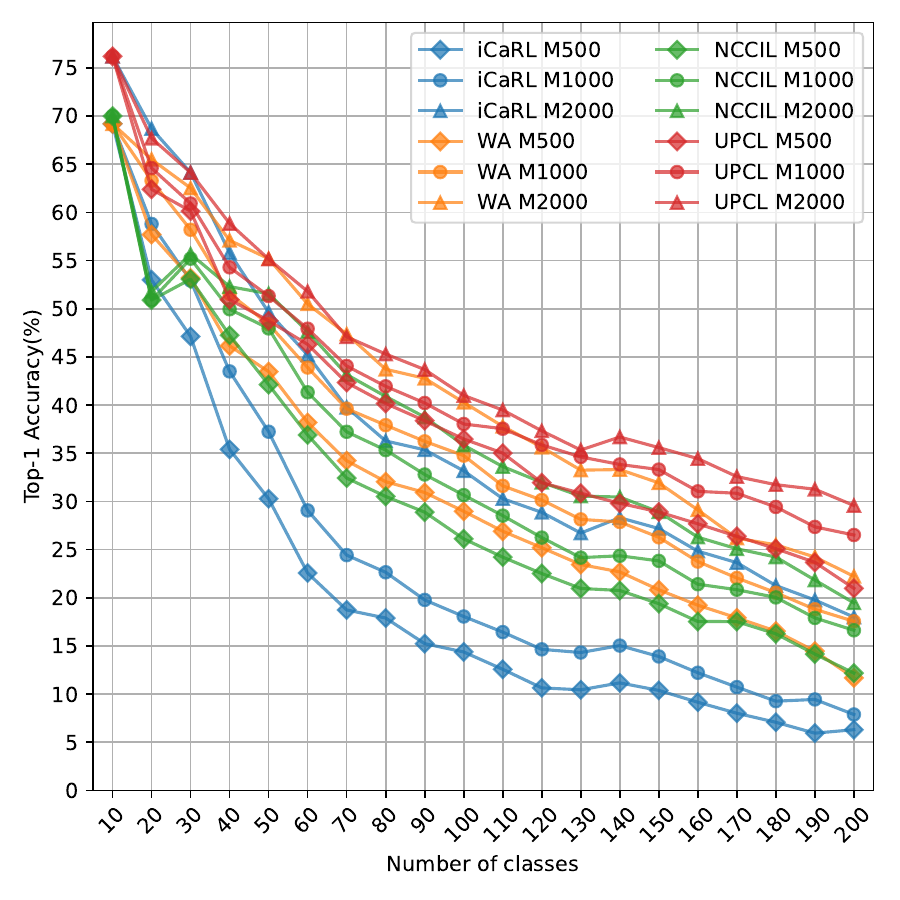}\label{subfig:tinyimagenet_0_10}}
    \caption{The performance with different memory sizes. M500, M1000, and M2000 represent the total number of exemplars in memory. }
\label{fig:memory}
\end{figure*}

\noindent \textbf{The performance with different memory sizes.} Fig.~\ref{fig:memory} shows the performance variation of both ImageNet100 and TinyImageNet with different memory sizes. And more detailed results of other task settings are presented in the supplementary material. 
$\Delta_{last}=\mathcal{A}_\text{last}^{M2000}-\mathcal{A}_\text{last}^{Mn}$ represents the performance fluctuations compared to M2000. The smaller $\Delta_{last}$ indicates that that the method is more susceptible to imbalances. And it indirectly means that the method is more difficult to deal with data imbalance. The results demonstrate that UPCL outperforms the other methods in terms of $\mathcal{A}_\text{last}$ and $\mathcal{A}_\text{avg}$. Among the compared methods, UPCL has the lowest $\Delta_{last}$, followed by NCCIL, which also uses fixed prototypes. This method of pre-allocating feature space can more effectively utilize a small amount of rehearsal data and effectively deal with the imbalance between old classes in memory and new classes in the task, avoiding feature confusion.

\subsection{Ablation studies of each component}
\label{sect:analysis}

\begin{table}
\centering
\caption{\textbf{Ablation studies of each component on CIFAR100}.}
\label{tab:ablation}
\scalebox{0.9}{
\begin{tblr}{
  column{1} = {r},
  column{2-7} = {c},
  hline{1,9} = {-}{0.1em},
  hline{2} = {-}{},
  vline{2,6} = {-}{},
}
                & {Cosine \\Classifier} & {Uniform \\Prototype} & {Fixed \\Margin} & {Dynamic \\Margin} & {\\ $\mathcal{A}_\text{last}$} & {\\ $\mathcal{A}_\text{avg}$} \\
Baseline (iCaRL) &                       &                       &                  &                    & 22.91                     & 43.98                    \\
Baseline+Cos       & \checkmark            &                       &                  &                    & 27.14                     & 44.83                    \\
Baseline+Cos+FM    & \checkmark            &                       & \checkmark       &                    & 28.03                     & 44.88                    \\
Baseline+Cos+DM    & \checkmark            &                       &                  & \checkmark         & 35.41                     & 49.78                    \\
Baseline+UP         &                       & \checkmark            &                  &                    & 40.62                     & 58.28                    \\
Baseline+UP+FM      &                       & \checkmark            & \checkmark       &                    & 41.72                     & 58.85                    \\
Baseline+UP+DM (\textbf{Ours})   &                       & \checkmark            &                  & \checkmark         & 51.78                     & 64.68                    
\end{tblr}
}
\end{table}


We conducted a series of ablation studies to delve into the effectiveness of the components of our proposed UPCL. The ablation experiments were conducted on CIFAR100 with a memory size of 500 and 10 new classes per task (T=10), and the remaining hyper-parameters are the same as those in the previous section. Specifically, we use iCaRL as the \textit{Baseline}. \textit{Baseline+Cos} uses a learnable cosine classifier and softmax cross-entropy loss function. \textit{Baseline+UP} uses the unlearnable uniform prototype as classifiers and prototype contrastive loss function. 

As can be observed from Table~\ref{tab:ablation}, despite using review and knowledge distillation, \textit{baseline+cos} still suffered from severe catastrophic forgetting, and the prediction results are biased towards the new classes. After adding the dynamic margin, the $\mathcal{A}_\text{last}$ and $\mathcal{A}_\text{avg}$ increased by 8.27\% and 4.95\% respectively, which indicates that the compact features and larger margin alleviated the confusion of features. After replacing the cosine classifier with the uniform prototype, the $\mathcal{A}_\text{last}$ and $\mathcal{A}_\text{avg}$ achieved a larger improvement, indicating that the uniform prototype can guide the backbone network to learn more discriminative features. When using the uniform prototype and dynamic margin at the same time, the model performance is further improved, indicating that these two components are independent and can work together to improve performance. 
We further compare our method with its fixed-margin counterpart to demonstrate the effectiveness of the proposed dynamic margin, and empirically set the margin to 0.1. 
In comparison to the baseline model, incorporating a fixed margin strategy does enhance performance. However, its effectiveness is inferior to that of the dynamic margin approach. Notably, the fixed margin introduces an additional hyper-parameter that necessitates fine-tuning based on data distribution. In contrast, the dynamic margin consistently outperforms the fixed margin, while also alleviating the burden of hyper-parameter search.

\section{Conclusion}

In this paper, we propose a novel class-incremental learning (CIL) method, uniform prototype contrastive learning (UPCL), from the perspective of imbalanced learning. UPCL learn class-uniform and feature-compact representations to improve the performance of CIL. Specifically, UPCL leverages a set of pre-defined uniform prototypes to guide the feature learning through contrastive learning. Moreover, UPCL dynamically adjusts the relative margin between old and new class features based on the prior distribution of the training data, which can balance the feature distribution and alleviate the catastrophic forgetting. We conduct extensive experiments on several benchmark datasets, including CIFAR100, ImageNet100 and TinyImageNet, and demonstrate that UPCL achieves state-of-the-art performance on CIL. However, the limitation of our method is that the number of classes supported cannot be greater than the dimensions of feature. Our future work will explore how to design prototypes with dynamic dimensionality and tackle the problem of unknown the total number of classes, thus further improve the performance of lifelong learning.

\bibliographystyle{unsrtnat}
\bibliography{main}


\newpage
\appendix

\section{Supplemental material}

\subsection{Total loss function}

To ensure a fair comparison, we follow a simple pipeline similar to iCaRL. Our method do not use any complex techniques such as dynamic networks, advanced rehearsal strategies, self-supervised learning auxiliary tasks, or hyper-parameter tuning tricks. For representation learning, we replace the basic KD in iCaRL with feature KD, which is defined as 
$$\mathcal{L}_{\text{FKD}} = \frac{1}{N} \sum_{0}^N 1 - {v_{tea}^i}^T \cdot {v_{stu}^i},$$
where $v_{tea}$ and $v_{stu}$ are the feature vectors of the teacher and student models, respectively.

In regards to the loss function $\mathcal{L}_{\text{UPCL}}$, we pragmatically decompose it into two components: $\mathcal{L}_{\text{feat}}$ and $\mathcal{L}_{\text{proto}}$ in implementation. 
$$\mathcal{L}_{\text{UPCL}} = \lambda_1\mathcal{L}_{\text{feat}}+\mathcal{L}_{\text{proto}},$$
we empirically set $\lambda_1 = 1/{2^t}$, where $t$ represents the number of tasks. 

Then, our total loss function is 
$$\mathcal{L} = (1-\lambda_2)\mathcal{L}_{\text{UPCL}} + \lambda_2\mathcal{L}_{\text{FKD}},$$
\noindent where $\lambda_2=C_{old}/C_{total}$, $C_{old}$ represents the number of old classes, and $C_{total}$ is the total number of classes.

\subsection{Optimizers and Hyperparameters}

To ensure a fair comparison, we have implemented the UPCL on the PyCIL framework. Additionally, several comparison methods, including iCaRL, LwF, WA, FeTrIL, and NCCIL, have been faithfully reproduced within this framework. Furthermore, other comparison methods, such as Coil, BiC, and ScrollNet, have been sourced from either the original paper’s presented results or the authors’ provided code reproductions.

All compared methods apply the same rehearsal strategy as iCaRL, and the memory size is set to 2000 unless otherwise specified. In all experiments, the methods use ResNet18 as the backbone network, and SGD optimizer. We do not use any tricks such as gridsearch to find the optimal hyper-parameters, but instead we use the default hyper-parameters in PyCIL for all methods in all experiments.  The detailed hyper-parameters are shown in the Table~\ref{tab:param}. All experiments are conducted on a single Nvidia RTX3090 GPU. The source code is provided in the Supplementary Material.

\begin{table}[!h]
\caption{\textbf{Hyperparameters}. We do not use any tricks such as gridsearch to find the optimal hyper-parameters, but instead we use the default hyper-parameters in PyCIL for all methods in all experiments. All our experiments are conducted on a single Nvidia RTX3090 GPU.}
\label{tab:param}
\centering
\begin{tblr}{
  cells = {c},
  cell{3}{2} = {c=2}{},
  cell{6}{2} = {c=2}{},
  hline{1,7} = {-}{0.08em},
  hline{2} = {-}{},
}
Hyperparameter & Base task & Increment task \\
Learning rate  & 0.1          & 0.1               \\
Weight decay   & 0.0002       &                   \\
Epochs         & 200          & 170               \\
Milestones     & 60,120,170   & 80,120,150        \\
Batch size     & 256          &                   
\end{tblr}

\end{table}

\subsection{Comprehensive Experimental Results}

In this section, we present detailed experimental results across three datasets: CIFAR100, Imagenet100, and TinyImagenet. We conducted multiple experiments with varying task numbers and memory sizes. Overall, our proposed approach achieves superior performance.

\begin{table}[!h]
\caption{\textbf{The experiment is conducted on CIFAR100, and T=5.} M500, M1000, and M2000 represent the total number of exemplars in memory. $\Delta$ represents the performance fluctuations compared to M2000.}
\label{tab:cifar_t5}
\centering
\scalebox{0.95}{
\begin{tblr}{
  width = \linewidth,
  colspec = {Q[119]Q[96]Q[96]Q[109]Q[96]Q[96]Q[109]Q[96]Q[96]},
  cells = {c},
  cell{1}{1} = {r=2}{},
  cell{1}{2} = {c=3}{0.301\linewidth},
  cell{1}{5} = {c=3}{0.301\linewidth},
  cell{1}{8} = {c=2}{0.192\linewidth},
  hline{1,7} = {-}{0.1em},
  hline{2} = {2-4}{leftpos = -1, rightpos = -1, endpos},
  hline{2} = {5-7}{leftpos = -1, rightpos = -1, endpos},
  hline{2} = {8-9}{leftpos = -1, rightpos = -1, endpos},
  hline{3} = {-}{},
}
      & M500  &       &        & M1000 &       &        & M2000 &       \\
      & $\mathcal{A}_\text{last}$  & $\mathcal{A}_\text{avg}$   & $\Delta_{last}$   & $\mathcal{A}_\text{last}$  & $\mathcal{A}_\text{avg}$   & $\Delta_{last}$   & $\mathcal{A}_\text{last}$  & $\mathcal{A}_\text{avg}$   \\
      iCaRL & 33.44 & 54.76 & -14.04 & 40.23 & 60.11 & -7.25 & 47.48 & 63.72\\
      WA & 54.28 & 66.01 & -8.99 & 59.38 & 69.80 & -3.72 & 63.27 & 72.13\\
      NCCIL & 47.52 & 56.83 & -6.23 & 49.58 & 58.20 & -4.07 & 53.65 & 61.09\\
      UPCL & \textbf{56.41} & \textbf{65.38} & \textbf{-3.91} & \textbf{58.88} & \textbf{67.39} & \textbf{-2.49} & \textbf{60.92} & \textbf{70.82}
\end{tblr}}
\end{table}

\begin{table}[!h]
\caption{\textbf{The experiment is conducted on CIFAR100, and T=10.} M500, M1000, and M2000 represent the total number of exemplars in memory. $\Delta$ represents the performance fluctuations compared to M2000.}
\label{tab:cifar_t10}
\centering
\scalebox{0.95}{
\begin{tblr}{
  width = \linewidth,
  colspec = {Q[119]Q[96]Q[96]Q[109]Q[96]Q[96]Q[109]Q[96]Q[96]},
  cells = {c},
  cell{1}{1} = {r=2}{},
  cell{1}{2} = {c=3}{0.301\linewidth},
  cell{1}{5} = {c=3}{0.301\linewidth},
  cell{1}{8} = {c=2}{0.192\linewidth},
  hline{1,7} = {-}{0.1em},
  hline{2} = {2-4}{leftpos = -1, rightpos = -1, endpos},
  hline{2} = {5-7}{leftpos = -1, rightpos = -1, endpos},
  hline{2} = {8-9}{leftpos = -1, rightpos = -1, endpos},
  hline{3} = {-}{},
}
      & M500  &       &        & M1000 &       &        & M2000 &       \\
      & $\mathcal{A}_\text{last}$  & $\mathcal{A}_\text{avg}$   & $\Delta_{last}$   & $\mathcal{A}_\text{last}$  & $\mathcal{A}_\text{avg}$   & $\Delta_{last}$   & $\mathcal{A}_\text{last}$  & $\mathcal{A}_\text{avg}$   \\
       iCaRL & 22.91 & 43.98 & -19.60 & 31.33 & 51.36 & -11.18 & 42.51 & 61.01\\
       WA & 42.45 & 56.34 & -13.46 & 48.56 & 61.91 & -7.35 & 55.91 & 68.22\\
       NCCIL & 37.17 & 52.91 & -11.36 & 42.37 & 56.04 & -6.16 & 48.53 & 60.93\\
       UPCL & \textbf{52.31} & \textbf{64.90} & \textbf{-4.34} & \textbf{52.87} & \textbf{66.74} & \textbf{-3.25} & \textbf{56.12} & \textbf{69.41}
\end{tblr}}
\end{table}

\begin{table}[!h]
\caption{\textbf{The experiment is conducted on CIFAR100, and T=20} M500, M1000, and M2000 represent the total number of exemplars in memory. $\Delta$ represents the performance fluctuations compared to M2000.}
\label{tab:cifar_t20}
\centering
\scalebox{0.95}{
\begin{tblr}{
  width = \linewidth,
  colspec = {Q[119]Q[96]Q[96]Q[109]Q[96]Q[96]Q[109]Q[96]Q[96]},
  cells = {c},
  cell{1}{1} = {r=2}{},
  cell{1}{2} = {c=3}{0.301\linewidth},
  cell{1}{5} = {c=3}{0.301\linewidth},
  cell{1}{8} = {c=2}{0.192\linewidth},
  hline{1,7} = {-}{0.1em},
  hline{2} = {2-4}{leftpos = -1, rightpos = -1, endpos},
  hline{2} = {5-7}{leftpos = -1, rightpos = -1, endpos},
  hline{2} = {8-9}{leftpos = -1, rightpos = -1, endpos},
  hline{3} = {-}{},
}
      & M500  &       &        & M1000 &       &        & M2000 &       \\
      & $\mathcal{A}_\text{last}$  & $\mathcal{A}_\text{avg}$   & $\Delta_{last}$   & $\mathcal{A}_\text{last}$  & $\mathcal{A}_\text{avg}$   & $\Delta_{last}$   & $\mathcal{A}_\text{last}$  & $\mathcal{A}_\text{avg}$   \\
       iCaRL & 18.84 & 39.43 & -22.75 & 26.00 & 48.46 & -15.59 & 41.59 & 48.46\\
       WA & 28.87 & 46.83 & -16.84 & 37.82 & 54.33 & -8.49 & 45.71 & 63.28\\
       NCCIL & 28.94 & 48.67 & -13.90 & 36.34 & 54.83 & -6.50 & 42.84 & 60.08\\
       UPCL & \textbf{40.41} & \textbf{57.89} & \textbf{-10.89} & \textbf{47.75} & \textbf{63.90} & \textbf{-3.55} & \textbf{51.30} & \textbf{67.16}
\end{tblr}}
\end{table}

\textbf{CIFAR100 Results (Tables \ref{tab:cifar_t5}, \ref{tab:cifar_t10}, \ref{tab:cifar_t20})}: We conducted experiments with task numbers of 5, 10, and 20 (corresponding to incremental class additions of 20, 10, and 5 per task). Notably, as the number of examples in memory decreases, the performance of various methods significantly declines. However, our method exhibits comparatively minimal performance degradation.

\begin{table}[!h]
\caption{\textbf{The experiment is conducted on Imagenet100, and T=5} M500, M1000, and M2000 represent the total number of exemplars in memory. $\Delta$ represents the performance fluctuations compared to M2000.}
\label{tab:imagenet100_t5}
\centering
\scalebox{0.95}{
\begin{tblr}{
  width = \linewidth,
  colspec = {Q[119]Q[96]Q[96]Q[109]Q[96]Q[96]Q[109]Q[96]Q[96]},
  cells = {c},
  cell{1}{1} = {r=2}{},
  cell{1}{2} = {c=3}{0.301\linewidth},
  cell{1}{5} = {c=3}{0.301\linewidth},
  cell{1}{8} = {c=2}{0.192\linewidth},
  hline{1,7} = {-}{0.1em},
  hline{2} = {2-4}{leftpos = -1, rightpos = -1, endpos},
  hline{2} = {5-7}{leftpos = -1, rightpos = -1, endpos},
  hline{2} = {8-9}{leftpos = -1, rightpos = -1, endpos},
  hline{3} = {-}{},
}
      & M500  &       &        & M1000 &       &        & M2000 &       \\
      & $\mathcal{A}_\text{last}$  & $\mathcal{A}_\text{avg}$   & $\Delta_{last}$   & $\mathcal{A}_\text{last}$  & $\mathcal{A}_\text{avg}$   & $\Delta_{last}$   & $\mathcal{A}_\text{last}$  & $\mathcal{A}_\text{avg}$   \\
      iCaRL & 34.74 & 57.06 & -15.96 & 41.10 & 62.19 & -9.06 & 50.70 & 68.16\\
      WA & 54.12 & 68.36 & -9.58 & 58.52 & 71.55 & -6.18 & 63.10 & 74.70\\
      NCCIL & 50.88 & 60.90 & -6.90 & 54.70 & 63.07 & -3.08 & 57.78 & 64.25\\
      UPCL & \textbf{58.90} & \textbf{71.15} & \textbf{-4.93} & \textbf{60.50} & \textbf{72.30} & \textbf{-3.40} & \textbf{63.90} & \textbf{74.83}
\end{tblr}}
\end{table}

\begin{table}[!h]
\caption{\textbf{The experiment is conducted on Imagenet100, and T=10.} M500, M1000, and M2000 represent the total number of exemplars in memory. $\Delta$ represents the performance fluctuations compared to M2000.}
\label{tab:imagenet100_t10}
\centering
\scalebox{0.95}{
\begin{tblr}{
  width = \linewidth,
  colspec = {Q[119]Q[96]Q[96]Q[109]Q[96]Q[96]Q[109]Q[96]Q[96]},
  cells = {c},
  cell{1}{1} = {r=2}{},
  cell{1}{2} = {c=3}{0.301\linewidth},
  cell{1}{5} = {c=3}{0.301\linewidth},
  cell{1}{8} = {c=2}{0.192\linewidth},
  hline{1,7} = {-}{0.1em},
  hline{2} = {2-4}{leftpos = -1, rightpos = -1, endpos},
  hline{2} = {5-7}{leftpos = -1, rightpos = -1, endpos},
  hline{2} = {8-9}{leftpos = -1, rightpos = -1, endpos},
  hline{3} = {-}{},
}
      & M500  &       &        & M1000 &       &        & M2000 &       \\
      & $\mathcal{A}_\text{last}$  & $\mathcal{A}_\text{avg}$   & $\Delta_{last}$   & $\mathcal{A}_\text{last}$  & $\mathcal{A}_\text{avg}$   & $\Delta_{last}$   & $\mathcal{A}_\text{last}$  & $\mathcal{A}_\text{avg}$   \\
    iCaRL & 23.14 & 47.20 & -18.68 & 32.22 & 54.35 & -9.60 & 41.82 & 62.02\\
    WA & 43.36 & 62.15 & -12.26 & 48.48 & 66.26 & -7.14 & 55.62 & 70.71\\
    NCCIL & 40.76 & 58.37 & -11.68 & 45.86 & 61.92 & -6.58 & 52.44 & 65.27\\
    UPCL & \textbf{48.74} & \textbf{64.35} & \textbf{-7.26} & \textbf{54.28} & \textbf{68.61} & \textbf{-1.72} & \textbf{56.00} & \textbf{72.05}
\end{tblr}}
\end{table}

\begin{table}[!h]
\caption{\textbf{The experiment is conducted on Imagenet100, and T=20} M500, M1000, and M2000 represent the total number of exemplars in memory. $\Delta$ represents the performance fluctuations compared to M2000.}
\label{tab:imagenet100_t20}
\centering
\scalebox{0.95}{
\begin{tblr}{
  width = \linewidth,
  colspec = {Q[119]Q[96]Q[96]Q[109]Q[96]Q[96]Q[109]Q[96]Q[96]},
  cells = {c},
  cell{1}{1} = {r=2}{},
  cell{1}{2} = {c=3}{0.301\linewidth},
  cell{1}{5} = {c=3}{0.301\linewidth},
  cell{1}{8} = {c=2}{0.192\linewidth},
  hline{1,7} = {-}{0.1em},
  hline{2} = {2-4}{leftpos = -1, rightpos = -1, endpos},
  hline{2} = {5-7}{leftpos = -1, rightpos = -1, endpos},
  hline{2} = {8-9}{leftpos = -1, rightpos = -1, endpos},
  hline{3} = {-}{},
}
      & M500  &       &        & M1000 &       &        & M2000 &       \\
      & $\mathcal{A}_\text{last}$  & $\mathcal{A}_\text{avg}$   & $\Delta_{last}$   & $\mathcal{A}_\text{last}$  & $\mathcal{A}_\text{avg}$   & $\Delta_{last}$   & $\mathcal{A}_\text{last}$  & $\mathcal{A}_\text{avg}$   \\
    iCaRL & 19.30 & 39.74 & -17.12 & 26.90 & 47.88 & -9.52 & 36.42 & 56.94 \\
    WA & 30.04 & 52.61 & -16.56 & 38.40 & 58.93 & -8.56 & 46.96 & 65.36 \\
    NCCIL & 30.02 & 50.53 & -14.5 & 37.58 & 56.24 & -6.94 & 44.52 & 61.61 \\
    UPCL & \textbf{39.90} & \textbf{60.77} & \textbf{-8.98} & \textbf{42.0} & \textbf{61.14} & \textbf{-6.88} & \textbf{48.88} & \textbf{67.16} \\
\end{tblr}}
\end{table}

\textbf{Imagenet100 Results (Tables \ref{tab:imagenet100_t5}, \ref{tab:imagenet100_t10}, \ref{tab:imagenet100_t20})}: In the Imagenet100 experiments, we again considered task numbers of 5, 10, and 20. Imagenet100, with its higher image resolution and greater demand for backbone network representation, presents a more challenging scenario. Remarkably, at T=5, the performance of various methods on Imagenet100 closely resembles that on CIFAR100. However, as T increases to 20, significant performance drops occur for most methods, whereas our approach maintains relatively high performance even under these conditions.

\begin{table}[!h]
\caption{\textbf{The experiment is conducted on TinyImagenet, and T=10.} M500, M1000, and M2000 represent the total number of exemplars in memory. $\Delta$ represents the performance fluctuations compared to M2000. }
\label{tab:tiny_t10}
\centering
\scalebox{0.95}{
\begin{tblr}{
  width = \linewidth,
  colspec = {Q[119]Q[96]Q[96]Q[109]Q[96]Q[96]Q[109]Q[96]Q[96]},
  cells = {c},
  cell{1}{1} = {r=2}{},
  cell{1}{2} = {c=3}{0.301\linewidth},
  cell{1}{5} = {c=3}{0.301\linewidth},
  cell{1}{8} = {c=2}{0.192\linewidth},
  hline{1,7} = {-}{0.1em},
  hline{2} = {2-4}{leftpos = -1, rightpos = -1, endpos},
  hline{2} = {5-7}{leftpos = -1, rightpos = -1, endpos},
  hline{2} = {8-9}{leftpos = -1, rightpos = -1, endpos},
  hline{3} = {-}{},
}
      & M500  &       &        & M1000 &       &        & M2000 &       \\
      & $\mathcal{A}_\text{last}$  & $\mathcal{A}_\text{avg}$   & $\Delta_{last}$   & $\mathcal{A}_\text{last}$  & $\mathcal{A}_\text{avg}$   & $\Delta_{last}$   & $\mathcal{A}_\text{last}$  & $\mathcal{A}_\text{avg}$   \\
       iCaRL & 9.88 & 29.30 & -14.36 & 11.98 & 32.56 & -12.41 & 24.24 & 43.39\\
       WA & 24.74 & 42.54 & -8.29 & 28.65 & \textbf{46.25} & -4.57 & 33.03 & \textbf{49.82}\\
       NCCIL & 22.75 & 37.81 & -4.56 & 25.46 & 39.60 & -2.02 & 27.31 & 41.48\\
       UPCL & \textbf{30.87} & \textbf{45.26} & \textbf{-2.91} & \textbf{32.39} & 45.25 & \textbf{-1.39} & \textbf{33.78} & 47.84
\end{tblr}}
\end{table}

\begin{table}[!h]
\caption{\textbf{The experiment is conducted on TinyImagenet, and T=20.} M500, M1000, and M2000 represent the total number of exemplars in memory. $\Delta$ represents the performance fluctuations compared to M2000.}
\label{tab:tiny_t20}
\centering
\scalebox{0.95}{
\begin{tblr}{
  width = \linewidth,
  colspec = {Q[119]Q[96]Q[96]Q[109]Q[96]Q[96]Q[109]Q[96]Q[96]},
  cells = {c},
  cell{1}{1} = {r=2}{},
  cell{1}{2} = {c=3}{0.301\linewidth},
  cell{1}{5} = {c=3}{0.301\linewidth},
  cell{1}{8} = {c=2}{0.192\linewidth},
  hline{1,7} = {-}{0.1em},
  hline{2} = {2-4}{leftpos = -1, rightpos = -1, endpos},
  hline{2} = {5-7}{leftpos = -1, rightpos = -1, endpos},
  hline{2} = {8-9}{leftpos = -1, rightpos = -1, endpos},
  hline{3} = {-}{},
}
      & M500  &       &        & M1000 &       &        & M2000 &       \\
      & $\mathcal{A}_\text{last}$  & $\mathcal{A}_\text{avg}$   & $\Delta_{last}$   & $\mathcal{A}_\text{last}$  & $\mathcal{A}_\text{avg}$   & $\Delta_{last}$   & $\mathcal{A}_\text{last}$  & $\mathcal{A}_\text{avg}$   \\
    iCaRL & 6.31 & 20.78 & -11.69 & 7.89 & 24.98 & -10.11 & 18.00 & 37.66\\
    WA & 11.69 & 31.65 & -10.54 & 17.56 & 36.49 & -5.11 & 22.23 & 41.67\\
    NCCIL & 12.18 & 30.19 & -7.29 & 16.63 & 33.77 & -2.84 & 19.47 & 38.00\\
    UPCL & \textbf{20.98} & \textbf{39.12} & \textbf{-8.60} & \textbf{26.52} & \textbf{42.00} & \textbf{-3.06} & \textbf{29.58} & \textbf{44.75}
\end{tblr}}
\end{table}

\begin{table}[!h]
\caption{\textbf{The experiment is conducted on TinyImagenet, and T=40.} M500, M1000, and M2000 represent the total number of exemplars in memory. $\Delta$ represents the performance fluctuations compared to M2000. }
\label{tab:tiny_t40}
\centering
\scalebox{0.95}{
\begin{tblr}{
  width = \linewidth,
  colspec = {Q[119]Q[96]Q[96]Q[109]Q[96]Q[96]Q[109]Q[96]Q[96]},
  cells = {c},
  cell{1}{1} = {r=2}{},
  cell{1}{2} = {c=3}{0.301\linewidth},
  cell{1}{5} = {c=3}{0.301\linewidth},
  cell{1}{8} = {c=2}{0.192\linewidth},
  hline{1,7} = {-}{0.1em},
  hline{2} = {2-4}{leftpos = -1, rightpos = -1, endpos},
  hline{2} = {5-7}{leftpos = -1, rightpos = -1, endpos},
  hline{2} = {8-9}{leftpos = -1, rightpos = -1, endpos},
  hline{3} = {-}{},
}
      & M500  &       &        & M1000 &       &        & M2000 &       \\
      & $\mathcal{A}_\text{last}$  & $\mathcal{A}_\text{avg}$   & $\Delta_{last}$   & $\mathcal{A}_\text{last}$  & $\mathcal{A}_\text{avg}$   & $\Delta_{last}$   & $\mathcal{A}_\text{last}$  & $\mathcal{A}_\text{avg}$   \\
    iCaRL & 3.99 & 16.65 & -12.63 & 7.15 & 21.89 & -9.47 & 16.62 & 35.01\\
    WA & 8.74 & 23.08 & -4.75 & 11.65 & 28.55 & -1.84 & 13.49 & 32.90\\
    NCCIL & 8.60 & 23.49 & -7.27 & 11.94 & 28.71 & -3.93 & 15.87 & 34.64\\
    UPCL & \textbf{15.99} & \textbf{32.98} & \textbf{-3.66} & \textbf{17.66} & \textbf{35.27} & \textbf{-1.65} & \textbf{19.65} & \textbf{37.31}
\end{tblr}}
\end{table}

\textbf{Imagenet100 Results (Tables \ref{tab:tiny_t10}, \ref{tab:tiny_t20}, \ref{tab:tiny_t40})}: In the TinyImagenet experiments, we maintained consistent class addition numbers of 20, 10, and 5 per task, while varying the total number of tasks (10, 20, and 40). TinyImagenet’s larger number of classes introduces additional complexity to CIL, exacerbating the catastrophic forgetting problem. In comparison to other methods, our approach demonstrates lighter catastrophic forgetting over longer sequences, resulting in superior performance.

\subsection{Ablation studies of each component}
\begin{figure}[htb]
\centering
    \subfloat[Baseline+Cos]{\includegraphics[width=.33\linewidth]{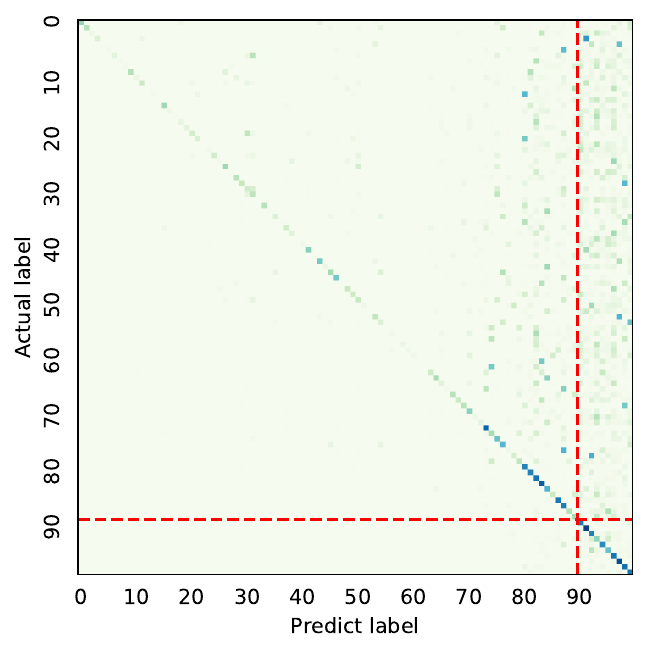}\label{subfig:baseline}}
    \subfloat[Baseline+UP]{\includegraphics[width=.33\linewidth]{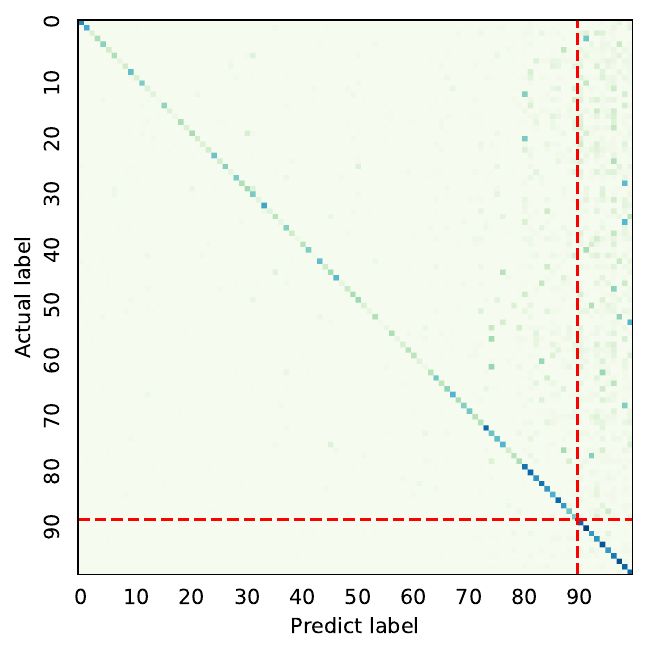}\label{subfig:up}}
    \subfloat[Baseline+Cos+FM]{\includegraphics[width=.33\linewidth]{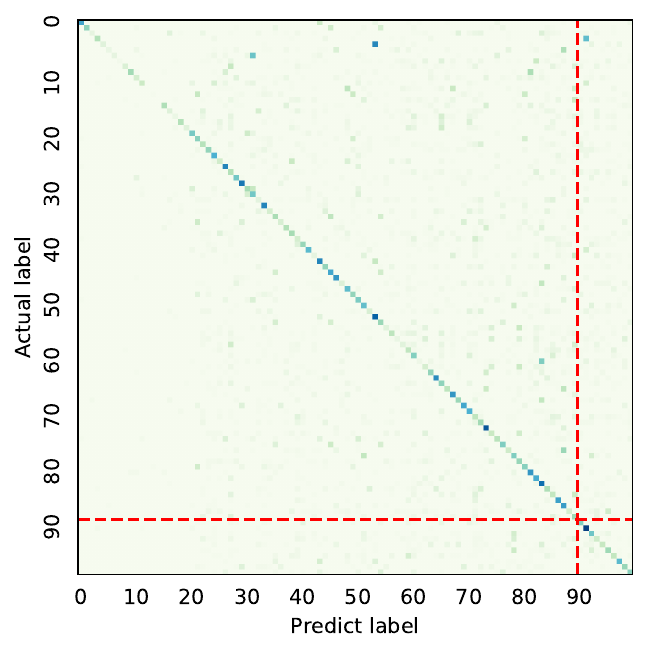}\label{subfig:baseline_fm}}\\
    \subfloat[Baseline+UP+FM]{\includegraphics[width=.33\linewidth]{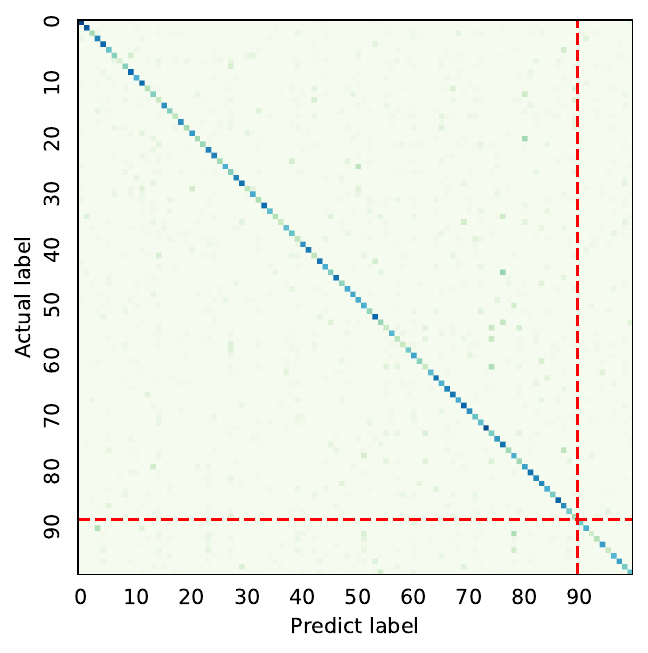}\label{subfig:up_fm}}
    \subfloat[Baseline+Cos+DM]{\includegraphics[width=.33\linewidth]{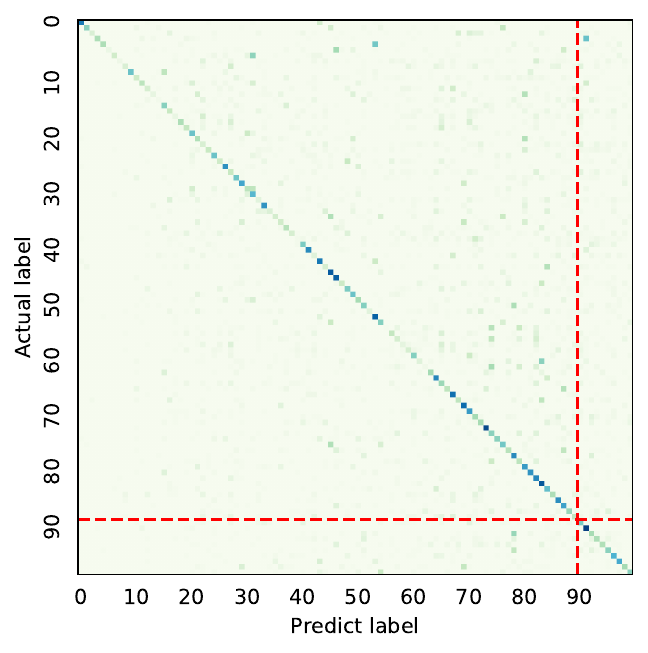}\label{subfig:baseline_dm}}
    \subfloat[Ours]{\includegraphics[width=.33\linewidth]{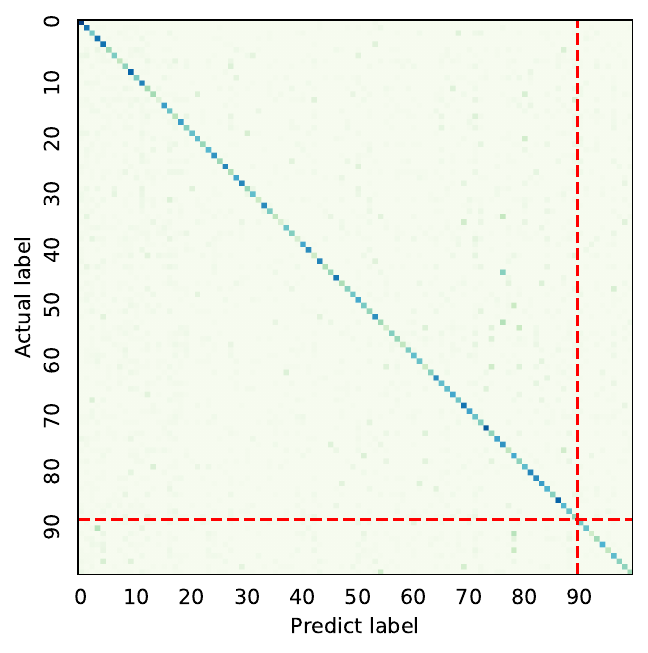}\label{subfig:up_dm}}
\caption{\textbf{Confusion matrices in ablation study of each component}.}
\label{fig:ablation_cm}
\end{figure}

Fig.~\ref{fig:ablation_cm} shows the confusion matrix. It can be seen that using margin alone avoids confusion between old and new classes, but it damages the learning of new classes. Using unlearnable uniform prototype alone alleviates the confusion between old and new classes. Using both components can significantly improve performance.

\subsection{Ablation studies of hyperparameter in the loss function}

\begin{table}[!h]
\caption{Impact of hyperparameter $\lambda_1$ on performance in the loss function.}
\label{tab:lambda}
\centering
\refstepcounter{table}
\begin{tblr}{
  width = \linewidth,
  hline{2} = {2-3}{leftpos = -1, rightpos = -1, endpos},
  hline{1,8} = {-}{0.1em},
}
 & $\mathcal{A}_\text{last}$ & $\mathcal{A}_\text{avg}$\\
$\lambda_1=0.1$ & \textit{51.02}& \textit{63.28}\\
$\lambda_1=1$ & 47.19 & 61.88\\
$\lambda_1=2$ & 44.86 & 59.91\\
$\lambda_1=1/10^t$ & 51.77 & 63.58\\
$\lambda_1=2^t$ & 10.94 & 45.83\\
$\lambda_1=1/2^t$ & \textbf{51.78} & \textbf{64.68}
\end{tblr}
\end{table}

In this experiment, we investigate the influence of the hyperparameter $\lambda_1$ within the loss function. The study is conducted on the CIFAR100 dataset, maintaining the same experimental setup as described in the ablation study section. As previously outlined in Section \ref{sec:loss}, we empirically set $\lambda_1=1/2^t$, where $t$ denotes the number of tasks.

The results in Table \ref{tab:lambda} demonstrate that directly summing the two loss components (i.e., $\lambda_1=1$) is not the most reasonable approach. During training, we observed that $\mathcal{L}_{\text{feat}}$ is typically 10 times larger than $\mathcal{L}_{\text{proto}}$. Consequently, an intuitive adjustment to bring them to the same order of magnitude is to set  $\lambda_1=0.1$. However, upon analyzing the loss function, we find that $\mathcal{L}_{\text{feat}}$ plays a crucial role in enabling the model to learn distinctive features during initial tasks. Yet, in subsequent tasks, it leads to overly uniform features, lacking compactness. To address this, we propose a dynamic hyperparameter that decreases with the progression of tasks: $\lambda_1=1/2^t$. From Table \ref{tab:lambda}, we observe that  $\lambda_1 < 1$ yield favorable results, while $\lambda_1 > 1$ severely compromise performance, thus reinforcing our hypothesis.

\begin{figure}[hbtp]
\centering
\begin{minipage}[t]{0.45\textwidth}
\centering
\includegraphics[width=\linewidth]{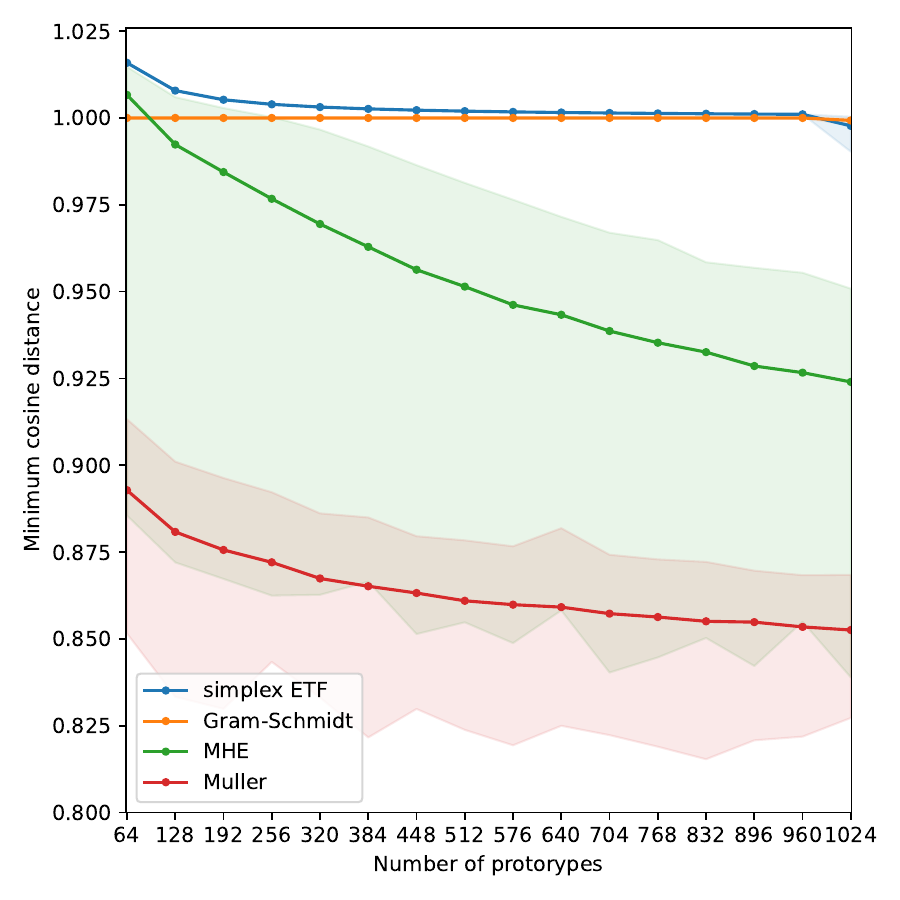}
\caption{Minimum cosine distance of prototypes generated by several methods. Larger values denotes better results.}
\label{fig:Uniformity} 
\end{minipage}
\hspace{10pt}
\begin{minipage}[t]{0.45\textwidth}
\centering
\includegraphics[width=\linewidth]{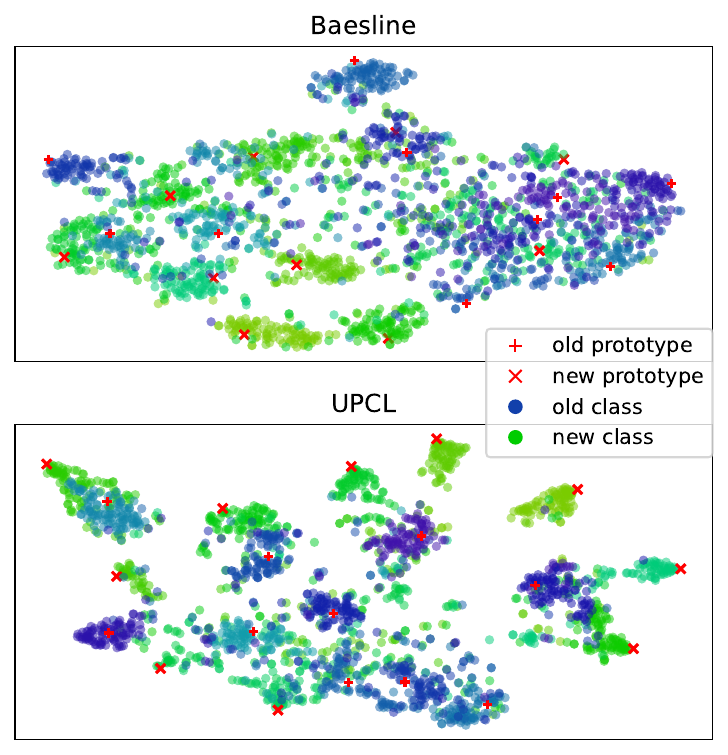}
\caption{The UPCL enhances the compactness of features within the same class while reducing overlap between old and new classes.}
\label{fig:vis}
\end{minipage}
\end{figure}

\subsection{Uniformity and orthogonality}

To validate the effectiveness of using orthogonal approximate uniformity, we compared the minimum cosine distances of prototypes generated by four methods: simplex ETF \cite{graf2021dissecting,papyan2020prevalence}, Gram-Schmidt algorithm \cite{cheney2009linear}, MHE \cite{liu2018learning}, and Muller’s procedure \cite{muller1959note}. Specifically, we set the feature dimension to 1024, and each method repeated 1000 times for each prototype quantity. Figure 6 shows the results of the experiment, where the shadow represents the range of the minimum cosine distance, and the line represents the mean. Simplex ETF represents the theoretical upper bound, and the effect of orthogonality is slightly lower than simplex ETF. The effect of MHE gradually decreases with the increase in the number of prototypes, which may be due to the need for more detailed search of training hyperparameters. Muller’s procedure does not require complex calculations, but its uniformity is lower than other methods. In summary, the result shows that orthogonality indeed approximates uniformity. 
When the number of classes is greater than the dimensionality, the embedding space should be increased, and designing prototypes with dynamic dimensionality will be our future work.

\subsection{Visualization}

We saved the features and prototypes of the baseline and UPCL in the last task of the ablation experiment, and visualized the final 20 classes using t-SNE. The blue points represents the old class features, the green points represents the new class features, and ‘+’ and ‘x’ represent the prototypes of the new and old classes, respectively. From Fig.~\ref{fig:vis}, it can be seen that compared with the baseline, UPCL can make the features more compact, thereby reducing the confusion between the new and old classes.



\end{document}